\pdfoutput=1

\documentclass[11pt]{article}

\usepackage[preprint]{acl}

\usepackage{times}
\usepackage{latexsym}

\usepackage[T1]{fontenc}

\usepackage[utf8]{inputenc}

\usepackage{microtype}

\usepackage{inconsolata}

\usepackage{graphicx}

\usepackage{setspace}
\usepackage{booktabs}
\usepackage{tabularx}
\usepackage{amssymb}
\usepackage{pifont}
\usepackage{xspace}
\usepackage{stfloats}
\usepackage{float}
\usepackage[most]{tcolorbox}
\usepackage{placeins}

\newcommand{\cmark}{\ding{51}}  

\newcommand{\easydataset}{Easy Dataset\xspace}
\newcommand{\llamafactory}{LlamaFactory\xspace}

\newsavebox{\myboxtable}

\title{\easydataset: A Unified and Extensible Framework for Synthesizing LLM Fine-Tuning Data from Unstructured Documents}

\author{
Ziyang Miao\textsuperscript{1},
Qiyu Sun\textsuperscript{1},
Jingyuan Wang\textsuperscript{1},
Yuchen Gong\textsuperscript{1}, \\
{\bf
Yaowei Zheng\textsuperscript{1},
Shiqi Li\textsuperscript{2}\thanks{Corresponding author: 1009903985@qq.com},
Richong Zhang\textsuperscript{1}\thanks{Corresponding author: zhangrc@act.buaa.edu.cn}
} \\
\textsuperscript{1}School of Computer Science and Engineering, Beihang University, China \\
\textsuperscript{2}Independent Researcher \\
Open-source repository: \textbf{\url{https://github.com/ConardLi/easy-dataset}} \\
Demonstration video: \textbf{\url{https://youtu.be/HlyvdE1ASRk}}
}

\begin{document}

\maketitle

\begin{abstract}

Large language models (LLMs) have shown impressive performance on general-purpose tasks, yet adapting them to specific domains remains challenging due to the scarcity of high-quality domain data. Existing data synthesis tools often struggle to extract reliable fine-tuning data from heterogeneous documents effectively. To address this limitation, we propose \easydataset, a unified framework for synthesizing fine-tuning data from unstructured documents via an intuitive graphical user interface (GUI). Specifically, \easydataset allows users to easily configure text extraction models and chunking strategies to transform raw documents into coherent text chunks. It then leverages a persona-driven prompting approach to generate diverse question-answer pairs using public-available LLMs. Throughout the pipeline, a human-in-the-loop visual interface facilitates the review and refinement of intermediate outputs to ensure data quality. Experiments on a financial question-answering task show that fine-tuning LLMs on the synthesized dataset significantly improves domain-specific performance while preserving general knowledge. The source code and installable package are available at \url{https://github.com/ConardLi/easy-dataset} and have garnered over 9,000 GitHub stars.

\end{abstract}

\section{Introduction}

Large language models (LLMs) have demonstrated remarkable capabilities across general-purpose applications \cite{achiam2023gpt,yang2025qwen3}, leading to their widespread deployment in both academia and industry. Nevertheless, as real-world applications increasingly demand domain-specific knowledge, effectively adapting LLMs to specialized scenarios remains a persistent challenge. Among existing approaches, supervised fine-tuning \cite{ouyang2022training,alpaca} has emerged as a practical and effective method for domain adaptation, enabling models to internalize domain-specific knowledge through learning from curated labeled data. Yet the effectiveness of fine-tuning depends on the availability of large-scale, high-quality datasets tailored to the target domain. In practice, constructing such datasets manually is often costly and time-consuming, particularly in domains that require expert annotations. As a result, automated data synthesis \cite{wang2024survey,tan2024large} has gained increasing attention as an attractive alternative to reduce human effort while preserving data diversity and domain relevance.

Despite the promise of automated data synthesis, implementing it in practice remains non-trivial due to two primary challenges: effectively parsing heterogeneous and noisy source documents and generating large-scale, high-quality supervised data. The first challenge stems from the structural complexity of real-world documents, which often contain a combination of unstructured and semi-structured elements, including free text, tables, and figures. As a result, standard parsing methods often struggle to handle this variability robustly and consistently \cite{zhang2024document}, resulting in incomplete or inaccurate information extraction. The second challenge involves generating diverse and faithful question-answer (QA) pairs that accurately capture the underlying domain knowledge. Simply reusing or repeating synthesized QA pairs can lead to overfitting and degrade downstream performance after fine-tuning. Therefore, an effective data augmentation strategy is crucial to generate varied examples while preserving semantic correctness and consistency with domain-specific content.

Recent studies have investigated automated data synthesis methods, such as constructing fine-tuning datasets based on predefined topics \cite{kiln} or problem sets \cite{distilabel-argilla-2024,curator2025}. However, these tools generally lack reliable capabilities for parsing unstructured documents to extract domain-specific data. \citet{djv1,djv2} introduce various document processing techniques, but they fall short of yielding high-quality QA pairs that are directly suitable for fine-tuning. Consequently, there remains a clear need for an end-to-end solution for synthesizing fine-tuning data from raw documents to effectively improve the performance of LLMs in domain-specific applications.

\begin{table*}[t]
\centering
\caption{Comparison of existing synthetic data generation tools.}
\label{tab:comparison}
\resizebox{.95\textwidth}{!}{%
\begin{tabular}{lcccccc}
\toprule
\textbf{Feature} & \textbf{Distilabel} & \textbf{Kiln} & \textbf{Curator} & \textbf{Data-Juicer} & \textbf{GraphGen} & \textbf{\easydataset} \\
\midrule
Adaptive Parsing              & \cmark & & \cmark & \cmark & \cmark & \cmark \\
Hybrid Chunking               &        & & \cmark & \cmark & \cmark & \cmark \\
Persona-Driven Data Synthesis &        & &        &        &        & \cmark \\
Human-In-The-Loop             &        & \cmark & &        &        & \cmark \\
End-to-End GUI                &        & \cmark & & \cmark & \cmark & \cmark \\
\bottomrule
\end{tabular}%
}
\end{table*}

We introduce \easydataset, a unified framework for synthesizing fine-tuning data from unstructured documents through an intuitive graphical user interface (GUI). It comprises two main components: {\bf adaptive document processing} and {\bf persona-driven data synthesis}. To transform raw documents into coherent text chunks, we integrate a suite of text extraction models, enabling robust parsing across various formats. The extracted textual content is subsequently segmented into semantically coherent chunks using a hybrid chunking strategy. To construct high-quality datasets from these chunks, we adopt a persona-driven prompting approach that generates diverse and large-scale question-answer (QA) pairs by leveraging publicly available LLMs. Importantly, the GUI facilitates human-in-the-loop refinement, allowing users to review and improve the quality of the synthesized data. It further allows users to execute the entire data synthesis workflow without writing any code, greatly enhancing usability for non-technical users.

Our contributions are summarized as follows:
\begin{itemize}
\item We introduce \easydataset, an extensible end-to-end framework that unifies adaptive document processing and persona-driven data synthesis to automate the curation of high-quality fine-tuning data from unstructured documents, while minimizing manual effort.
\item We develop an intuitive visual interface that enhances accessibility for non-technical users and facilitates efficient human-in-the-loop refinement to ensure data quality.
\item Empirical results on a financial QA task show that fine-tuning LLMs on the synthesized data significantly improves domain-specific performance while retaining general knowledge.
\end{itemize}

\section{Related Work}

In recent years, numerous systems have been proposed to facilitate automated synthetic data generation. Distilabel~\cite{distilabel-argilla-2024} provides a modular framework with practical components for generating various types of synthetic data, including instruction-tuning and preference datasets. Kiln~\cite{kiln} features a GUI-based platform that supports zero-shot data generation, topic tree construction, and reasoning data creation, along with an interactive interface for human review and rating. Curator~\cite{curator2025} facilitates the synthesis of diverse data types, including reasoning, code execution, chart generation, and function-calling data, to support fine-tuning across a broad spectrum of downstream tasks. Data-Juicer~\cite{djv1,djv2} supports multiple user interfaces and provides a rich set of data processing operators, making it highly customizable. GraphGen~\cite{chen2025graphgen} leverages knowledge graphs to generate question-answer pairs from source documents through a visual interface. Despite their contributions, these tools still exhibit several notable limitations (Table~\ref{tab:comparison}). Many lack support for human-in-the-loop interaction, hindering opportunities for iterative quality refinement and alignment with user preferences. Others do not support synthesizing fine-tuning data based on source documents, making it difficult to efficiently construct domain-specific datasets. Moreover, few provide a fully end-to-end graphical user interface (GUI) that integrates the entire data synthesis pipeline, resulting in steep learning curves and limited usability for users with diverse technical backgrounds. These gaps underscore the need for a unified, accessible solution.

\section{Framework}

\begin{figure*}[t]
\centering
\includegraphics[width=.98\textwidth]{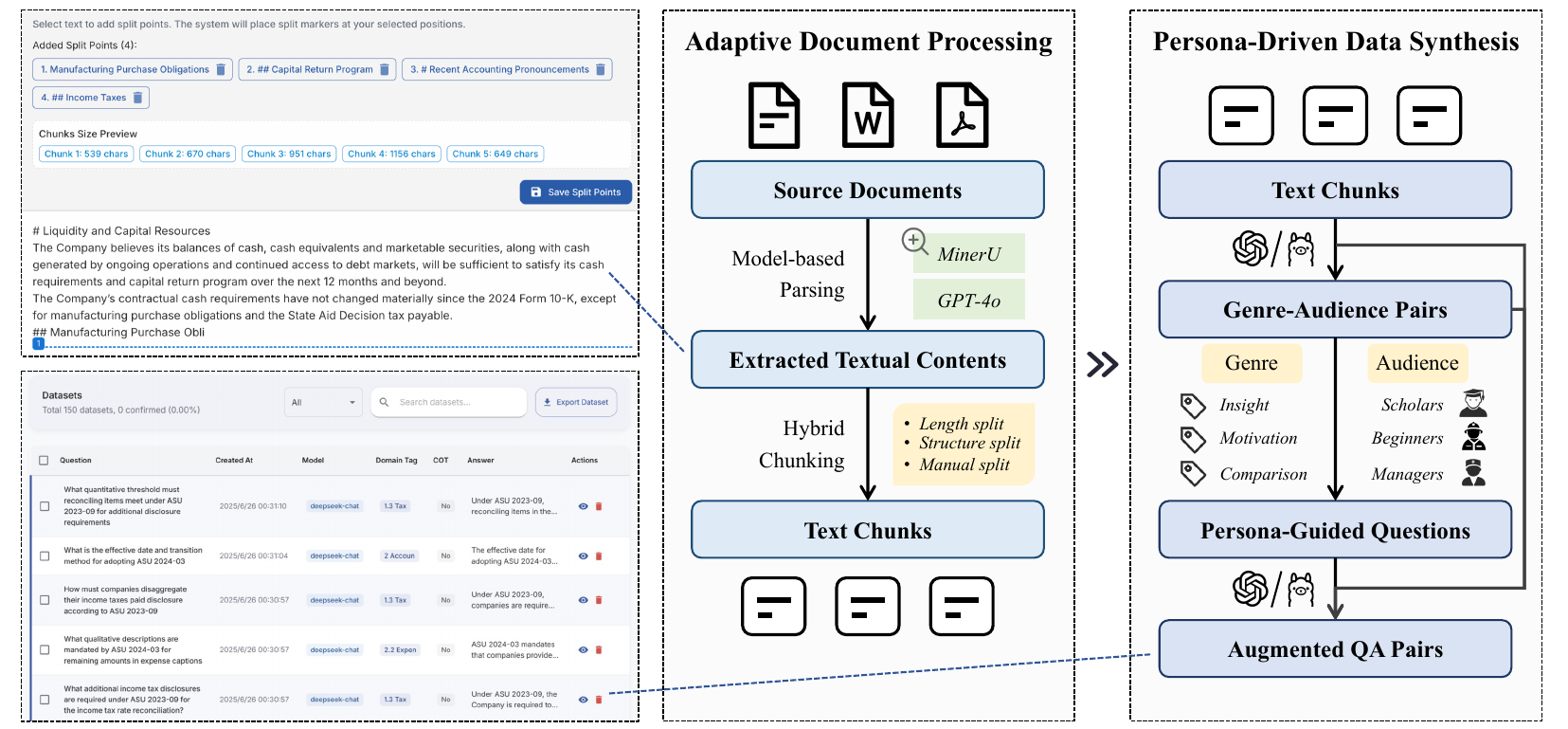}
\caption{
Overview of the \easydataset framework. The framework consists of two primary components: adaptive document processing and persona-driven data synthesis. In the first component, documents of different formats are processed via model-based parsing, followed by hybrid chunking to produce text chunks. In the second component, {\em Genre}-{\em Audience} pairs are created for each document to guide the construction of diverse question-answer pairs. Persona-guided questions are then generated to further increase diversity, and final augmented QA pairs are synthesized via knowledge-enhanced prompting to ensure factual consistency. The entire framework provides interactive human-in-the-loop refinement to maintain data quality.
}
\label{fig:framework}
\end{figure*}

We present \easydataset, a unified and extensible framework designed to synthesize high-quality fine-tuning datasets from unstructured documents. As illustrated in Figure~\ref{fig:framework}, the framework implements a visual human-in-the-loop pipeline that enables interactive refinement at all stages of dataset construction. It begins with adaptive document processing, leveraging vision-language models (VLMs) to extract textual content from raw documents. The extracted text is then segmented using a hybrid chunking strategy that combines length-based, structure-based, and manual splitting. In the QA generation stage, \easydataset supports both naive and persona-driven data synthesis to generate stylistically diverse and semantically grounded QA pairs. A modular model configuration interface enables flexible access to both API-based and locally deployed LLMs and offers fine-grained control over generation parameters. Finally, the framework supports export of datasets in multiple formats and is natively compatible with fine-tuning frameworks such as \llamafactory~\cite{zheng-etal-2024-llamafactory}.

\subsection{Adaptive Document Processing}

\paragraph{Document Parsing} Our framework supports extracting textual content from unstructured documents of various formats, including plain text, Markdown, DOCX, and PDF, which is essential for constructing high-quality datasets for downstream tasks. For plain text and Markdown documents, the original content is retained with minimal processing to preserve inherent semantics. In contrast, DOCX documents often lack explicit structure, necessitating preprocessing to extract meaningful text. To this end, we employ the lightweight Mammoth library \cite{mammoth} to convert DOCX files into Markdown, which helps preserve the original semantics while avoiding unnecessary formatting noise. For PDF documents with relatively simple layouts, we use the pdf2md tool \cite{pdf2md} to extract textual content efficiently. However, for PDFs with complex or mixed text-image compositions, rule-based methods often result in significant information loss. To address this limitation, we first apply layout analysis to automatically detect content regions within the document. Once the layout structure is established, textual regions are extracted directly, while visual regions are parsed using a vision-language model (VLM), which can accurately capture the content. This strategy ensures accurate text extraction and preserves key information, providing a reliable foundation for subsequent downstream data processing. In addition, our framework integrates with state-of-the-art PDF processing tools, including MinerU \cite{wang2024mineru,he2024opendatalab}, providing users with greater flexibility to handle documents of varying complexity and scale.

\paragraph{Text Chunking} To enable precise control over data sampling and ensure compatibility with existing LLM context windows, we propose {\em Hybrid-Chunking}, a structure-aware and adaptive preprocessing strategy that flexibly segments source documents into coherent text chunks. It allows users to configure chunk sizes and customize text delimiters, enabling adaptation to various content types, such as plain text, code snippets, or tabular data. The process begins with initial coarse-grained segmentation based on line breaks, followed by a hybrid splitting-and-merging routine: lengthy chunks are recursively split using user-defined delimiters, while neighboring short segments are merged if necessary to meet length constraints without breaking semantic units. Furthermore, for edge cases where automated rules may fail to work optimally, we provide a visual text-chunking interface that enables users to perform fine-grained manual adjustments, ensuring precise text segmentation. Users can flexibly tailor chunking policies and thresholds according to specific document content. This hybrid design achieves a balance between automation and user control, significantly improving the consistency and reliability of the resulting text chunks.

\subsection{Persona-Driven Data Synthesis}

\easydataset provides a flexible pipeline for generating question-answer (QA) pairs from text chunks. The process is highly configurable, supporting two generation modes: naive QA-pair generation and persona-driven QA-pair generation, the latter enabling the synthesis of diverse and stylistically controlled fine-tuning data.

\paragraph{Question Generation} Based on the text chunks, questions are automatically constructed by prompting the LLMs. Specifically, each text chunk is combined with a customizable system prompt before being passed to the LLM for question generation. This system prompt enables fine-grained control over question style, target audience, and tone. For example, it can specify whether questions should be concise, elaborative, or directive in nature. To further improve robustness in real-world scenarios, we introduce a stochastic punctuation dropout mechanism. In particular, question marks are randomly removed to prevent the model from over-relying on punctuation cues, thereby enhancing the generalization performance of models fine-tuned on the synthesized QA dataset.

\begin{table*}[t]
\centering
\caption{Persona-driven QA pairs synthesized by \easydataset. By varying the GA pair, our framework generates a diverse set of data examples that maintain high fidelity to the source documents.}
\label{tab:ga_examples}
\resizebox{.98\textwidth}{!}{%
\begin{tabular}{p{60em}}
\toprule
\textbf{Example 1: Financial News Summary for Executives and Business Readers} \\
\midrule
\textbf{Genre:} \textit{Financial News Summary.} This genre condenses complex financial information into concise, accessible updates. It highlights key points like liquidity status, capital returns, and regulatory changes, making it suitable for readers who need quick, high-level insights without deep technical details. \\
\addlinespace
\textbf{Audience:} \textit{Busy Executives and General Business Readers.} Professionals who need to stay informed about financial developments but lack the time for in-depth analysis. They prefer succinct summaries that provide the essential takeaways without overwhelming detail. \\
\midrule
\textbf{Generated Question:} \\
\textit{When is the Company expected to adopt ASU 2023-09, and what transition method will be used?} \\
\addlinespace
\textbf{Generated Answer:} \\
The Company will adopt ASU 2023-09 in its fourth quarter of 2026 using a prospective transition method. This update will enhance income tax disclosures by requiring additional details in the tax rate reconciliation and disaggregating taxes paid by federal, state, and foreign jurisdictions. \\
\midrule[1.5pt]
\textbf{Example 2: In-depth Policy Discussion for Professionals and Policy Makers} \\
\midrule
\textbf{Genre:} \textit{Tax and Accounting Policy Discussion.} This genre focuses on the implications of new tax and accounting policies, such as FASB updates. It explores how these changes affect corporate reporting, tax strategies, and financial transparency, often inviting debate and expert opinions. \\
\addlinespace
\textbf{Audience:} \textit{Tax Professionals and Policy Makers.} Experts in tax law and accounting standards who need to understand the practical and strategic impacts of regulatory changes. They are interested in how policies like ASU 2023-09 influence corporate behavior and financial disclosures. \\
\midrule
\textbf{Generated Question:} \\
\textit{How does ASU 2023-09 require companies to disaggregate income taxes paid by jurisdiction?} \\
\addlinespace
\textbf{Generated Answer:} \\
ASU 2023-09 mandates that companies disaggregate their income taxes paid disclosure by federal, state, and foreign taxes. Additionally, for foreign taxes, further disaggregation is required for significant individual jurisdictions. This enhanced disclosure framework aims to provide greater transparency into the geographic distribution of tax payments, enabling stakeholders to better assess the company's tax profile and jurisdictional exposures. {\it (truncated)} \\
\bottomrule
\end{tabular}%
}
\end{table*}

\paragraph{Answer Generation} Based on the generated questions, we derive high-quality answers that remain faithful to the source content. We employ a knowledge-enhanced prompting strategy, where the prompt contains the question and its corresponding source content. This ensures that the LLM-generated answers remain semantically aligned with the source content, factually consistent, and relevant to the overall domain context. The generation style is configurable to meet task-specific requirements, such as producing concise, detailed, or explanatory answers, allowing users to tailor the output style and format as needed. When using reasoning LLMs with chain-of-thought (CoT) capabilities, such as DeepSeek-R1~\cite{guo2025deepseek}, the intermediate reasoning steps are also included in the QA pairs to provide transparency and support more interpretable fine-tuning. This substantially improves interpretability and facilitates downstream error analysis. To ensure the quality and reliability of the generated QA pairs, we provide a post-generation refinement interface that allows users to manually review, edit, and verify answers. In addition, LLMs are used to automatically refine both final answers and their corresponding CoT reasoning traces, serving as an additional step to improve overall robustness and factual accuracy.

\paragraph{Persona-Driven QA-Pair Generation} \citet{ge2024scaling} first introduced the persona-driven data synthesis approach, demonstrating that prompting LLMs with different communicative roles (personas) enables the generation of stylistically diverse outputs, which can significantly increase dataset size. Building on this insight, we integrate persona-driven data synthesis into our \easydataset framework via a two-stage pipeline inspired by the Massive Genre-Audience (MGA) method \citep{hao2025reformulation}, which provides a systematic way to diversify output styles. Specifically, the pipeline first automatically generates diverse Genre-Audience (GA) pairs to guide stylistic conditioning and then produces multiple QA pairs conditioned on these personas, enriching the dataset with varied perspectives and tones. While MGA was originally developed to expand pretraining corpora by reformulating documents and reducing data redundancy, we adapt its core principles for QA pair synthesis in \easydataset to boost both the diversity and scale of the generated data for domain-specific fine-tuning.

Our adapted persona-driven pipeline consists of two stages for synthesizing diverse, persona-guided QA pairs from unstructured documents. In the {\bf Persona Synthesis Stage}, for each source document, we first leverage an LLM to automatically generate a set of unique ({\em Genre}, {\em Audience}) pairs based on the content. Each pair jointly defines a persona for the generation task. Specifically, {\em Genre} characterizes the inquisitive intent and dialogue style, while {\em Audience} profiles the cognitive state and knowledge background of the questioner. For example, a (Motivation, Beginners) persona guides the model to produce simple, encouragement-oriented questions that help novices build confidence. By introducing diverse Genre-Audience combinations, the synthesis process enables more comprehensive coverage of the original content from multiple perspectives. To enhance flexibility, users can also manually specify or refine GA pairs to better target specific domains or tasks. In the {\bf Persona-Guided QA Generation Stage}, the synthesized personas guide the LLMs to generate diverse questions based on the text chunks from various perspectives. For each generated question, the model subsequently produces an answer conditioned on the question, the corresponding source text chunk, and the associated persona. This yields an augmented QA pair in which both the question and the answer are semantically grounded in the original content and stylistically consistent with the intended persona. This adaptation effectively transforms the original MGA method from a pretraining augmentation strategy into a powerful framework for synthesizing diverse, persona-guided fine-tuning data. As illustrated in Table~\ref{tab:ga_examples}, this persona-driven approach enables the synthesis of diverse and faithful QA pairs from a single source document, thereby enhancing the effective utilization of the original source corpora.

\subsection{Model Configuration}

To seamlessly integrate LLMs into the data generation pipeline, we propose a flexible model configuration module. This module allows users to configure and manage multiple LLMs for data synthesis through an intuitive visual interface with minimal setup effort. By specifying the model provider, API endpoint, API key, and model name, users can easily utilize their target models. The module also supports locally deployed models via Ollama \cite{ollama2023}, ensuring flexibility for different deployment scenarios. Additionally, it provides fine-grained control over generation parameters (e.g., temperature and top-p sampling), enabling users to adapt the generation process to diverse data requirements and domain-specific constraints.

\subsection{Dataset Export}

\easydataset enables export of the generated QA pairs as standard dataset files in formats such as JSON, JSONL, and CSV. It also supports widely adopted data schemas such as Alpaca \cite{alpaca} and ShareGPT \cite{sharegpt}. Users can define custom export templates by specifying key fields such as question, answer, reasoning steps, and domain labels, allowing them to flexibly adapt outputs to diverse task-specific fine-tuning pipelines and community standards. Additionally, \easydataset offers seamless integration with the \llamafactory training framework \cite{zheng-etal-2024-llamafactory} by automatically generating a compatible configuration file. Users can simply specify the path to this configuration file to enable direct use in \llamafactory, eliminating manual setup and significantly lowering the barriers to fine-tuning.

\begin{table*}[t]
\centering
\caption{
Performance comparison of the Qwen2.5-7B-Instruct model before and after fine-tuning with naive and persona-driven data synthesis on general-purpose and domain-specific benchmarks. \textbf{Bold} indicates the best result, and \underline{underline} indicates the second-best.}
\label{tab:result}
\resizebox{.95\textwidth}{!}{%
\begin{tabular}{l|cccccc|c}
\toprule
Method & MMLU & CMMLU & HellaSwag & MATH & HumanEval & Avg. & Domain Knowledge \\
\midrule
Base Model & \underline{62.0} & 77.2 & \textbf{82.9} & \textbf{74.8} & \textbf{84.8} & \textbf{76.3} & 3.2 \\
Naive Data Synthesis & 60.2 & \textbf{78.4} & \underline{80.8} & 72.9 & 80.5 & 74.6 & \underline{57.0} \\
Persona-Driven Data Synthesis & \textbf{63.7} & \underline{77.7} & \underline{80.8} & \underline{74.1} & \underline{81.1} & \underline{75.5} & \textbf{59.6} \\
\bottomrule
\end{tabular}%
}
\end{table*}

\section{Evaluation}

\subsection{Experimental Setup}

To evaluate the quality of data synthesized by \easydataset, we collected five up-to-date (later than the knowledge cutoff) financial reports from public online sources as a representative domain example. We also built a domain-specific evaluation dataset consisting of 100 questions derived from the source documents. Then we used \easydataset to generate a training dataset and fine-tuned the Qwen2.5-7B-Instruct~\cite{qwen2024qwen25} model using the \llamafactory framework. We compared the model's performance before and after fine-tuning using either the naive or persona-driven data synthesis pipeline. We evaluated the model on the domain-specific dataset and general-purpose benchmarks, including MMLU~\cite{mmlu}, CMMLU~\cite{li-etal-2024-cmmlu}, HellaSwag~\cite{zellers-etal-2019-hellaswag}, MATH~\cite{math}, and HumanEval~\cite{humaneval}. For domain-specific evaluation, we employed the LLM-as-a-judge approach \cite{sharegpt} using the DeepSeek-V3 API \cite{liu2024deepseek} to provide reliable scores, with details provided in Appendix~\ref{app:eval}. Other fine-tuning details are provided in Appendix~\ref{app:detail}.

\subsection{Results}

Table~\ref{tab:result} shows the evaluation results of Qwen2.5-7B-Instruct on several general-purpose benchmarks and the domain-specific evaluation dataset, comparing the base model with its fine-tuned variants trained using either the naive or persona-driven data synthesis methods. On general-purpose benchmarks, we observe that fine-tuning on domain-specific data preserves the model's general language capabilities, demonstrating that the synthetic dataset produced by \easydataset enables the model to acquire domain-specific knowledge while maintaining robust generalization. Notably, the persona-driven variant achieves the best results on the MMLU benchmark and competitive performance across most tasks, even outperforming the naive variant on several challenging benchmarks, indicating its potential for improving generalization through stylistic and semantic diversity. On the domain-specific evaluation dataset, both fine-tuned models demonstrate substantial gains over the base model, which performs poorly without exposure to up-to-date financial documents. Specifically, fine-tuning with the dataset generated by \easydataset using the naive synthesis pipeline achieves a score of 57.0, while incorporating persona-driven data synthesis further improves the score to 59.6. These results highlight the effectiveness of \easydataset in injecting timely, domain-specific knowledge while preserving the model's general capabilities.

\section{Conclusion and Future Work}

We introduced \easydataset, an end-to-end framework that automates the synthesis of domain-specific fine-tuning datasets. With human-in-the-loop control and an intuitive interface, it enables efficient generation of datasets with high quality, diversity, and usability. Empirical results demonstrate that \easydataset significantly improves domain adaptation for LLMs while preserving general capabilities and robustness. We plan to extend \easydataset in several directions, including supporting broader modalities (e.g., SQL, tables, multi-modal), integrating automatic quality monitoring, and developing advanced enrichment strategies to further enhance data variety and fidelity.

\bibliography{main}
\clearpage

\appendix

\section{LLM-as-a-Judge Prompt}\label{app:eval}

\begin{tcolorbox}[colback=gray!10!white,
                  colframe=black,
                  boxrule=0.5pt,
                  arc=1pt,
                  left=6pt,right=6pt,
                  top=6pt,bottom=48pt]

Please act as an impartial evaluator and assess the quality of the AI assistant's response to the user's question. You will be provided with the following information:\\\\
1. The original user question (Question)  \\
2. A standard answer containing information directly relevant to the user's question (Ground Truth)  \\
3. The AI assistant's response (Prediction)  \\

Your task is to conduct a thorough evaluation focusing on correctness, scoring from 0 to 5 points. \\\\
\textbf{Evaluation Method:} \\
1. Carefully read the question, the assistant's response, and the ground truth answer.\\  
2. Identify and list all key factual statements from the ground truth.  \\
3. For each fact, determine whether it is correctly reflected in the assistant's response. \\  
4. Assign a final correctness score based on the degree of fact matching. If all facts from the ground truth are correctly reflected in the AI response, assign 5 points. If none are correct, assign 0 points. \\\\
Please carefully analyze the correctness of the answer. Finally, provide the score result in JSON format as follows:
\begin{verbatim}
[
  {
     "correctness": "3"
  }
]
\end{verbatim}

\noindent\textbf{Question} \\
\texttt{{\{ Question \}}} \\[0.5em]

\noindent\textbf{Prediction} \\
\texttt{{\{ Prediction \}}} \\[0.5em]

\noindent\textbf{Ground Truth} \\
\texttt{{\{ Ground Truth \}}}

\end{tcolorbox}

\section{Fine-Tuning Details}\label{app:detail}

All the experiments were conducted on 2 NVIDIA A800 80GB GPUs. For all fine-tuning runs, we used a batch size of 64 and a total of 10,000 training samples. The model was trained for 2 epochs with a learning rate of $10^{-5}$ using a cosine learning rate schedule. A warmup ratio of 0.1 was applied to stabilize training and mitigate overfitting.

\end{document}